\pdfoutput=1

\documentclass[11pt]{article}
\usepackage[]{acl}
\usepackage{graphicx}
\usepackage{times}
\usepackage{latexsym}
\usepackage{xspace}
\usepackage{stfloats}

\usepackage[T1]{fontenc}

\usepackage[utf8]{inputenc}

\usepackage{microtype}

\usepackage{algorithm}
\usepackage{algorithmic}
\usepackage{amsmath}
\usepackage{booktabs}
\usepackage{todonotes}
\usepackage{xspace}
\usepackage{caption}
\usepackage{subcaption}
\usepackage{hyperref}
\usepackage{cleveref}
\usepackage{xcolor}
\usepackage{amssymb}

\newcommand{\ours}{CeMAT\xspace}

\usepackage[markup=underlined]{changes}

\newcommand{\ANMT}{Autoregressive Neural Machine Translation\xspace}
\newcommand{\NAT}{Non-autoregressive Neural Machine Translation\xspace}
\newcommand{\subANMT}{AT\xspace}
\newcommand{\subNAT}{NAT\xspace}
\newcommand{\aligncs}{\emph{aligned code-switching \& masking}\xspace}
\newcommand{\dymask}{\emph{dynamic dual-masking}\xspace}

\newcommand{\undirec}{$\to$}
\newcommand{\bidirec}{$\leftrightarrow$}

\newcommand{\ANAT}{Autoregressive and Non-autoregressive Neural Machine Translation\xspace}
\newcommand{\exis}{$\bullet$}
\newcommand{\noexis}{$\ $}
\title{Universal Conditional Masked Language Pre-training \\ for Neural Machine Translation}

\newcommand*{\affmark}[1][*]{\textsuperscript{#1}}

\author{
  Pengfei Li\affmark[1]  \qquad Liangyou Li\affmark[1] \qquad Meng Zhang\affmark[1] \qquad Minghao Wu\affmark[2] \qquad Qun Liu\affmark[1] \\
  \affmark[1]Huawei Noah's Ark Lab \quad \affmark[2]Monash University \\
  \texttt{\{lipengfei111, liliangyou, zhangmeng92, qun.liu\}@huawei.com} \\
  \texttt{minghao.wu@monash.edu}
}

\begin{document}
\maketitle
\begin{abstract}

Pre-trained sequence-to-sequence models have significantly improved Neural Machine Translation (NMT). 
Different from prior works where pre-trained models usually adopt an unidirectional decoder, this paper demonstrates that pre-training a sequence-to-sequence model but with a bidirectional decoder can produce notable performance gains for both Autoregressive and Non-autoregressive NMT.
Specifically, we propose \ours, a conditional masked language model pre-trained on large-scale bilingual and monolingual corpora in many languages.\footnote{Code, data, and pre-trained models are available at \url{https://github.com/huawei-noah/Pretrained-Language-Model/tree/master/CeMAT}} 
We also introduce two simple but effective methods to enhance the \ours, \aligncs and \dymask. We conduct extensive experiments and show that our \ours can achieve significant performance improvement for all scenarios from low- to extremely high-resource languages, i.e., up to +14.4 BLEU on low-resource and +7.9 BLEU on average for Autoregressive NMT. 
For Non-autoregressive NMT, we demonstrate it can also produce consistent performance gains, i.e., up to +5.3 BLEU. To the best of our knowledge, this is the first work to pre-train a unified model for fine-tuning on both NMT tasks.


\end{abstract}

\section{Introduction}

Pre-trained language models have been widely adopted in NLP tasks \citep{DBLP:conf/naacl/DevlinCLT19,Radford2018ImprovingLU}. For example, XLM \citep{DBLP:conf/nips/ConneauL19} demonstrated that cross-lingual pre-training is effective in improving neural machine translation (NMT), especially on low-resource languages. These methods all directly pre-train a bidirectional encoder or an unidirectional decoder. The encoder and decoder in NMT models are then independently initialized with them and fine-tuned \citep{DBLP:conf/nips/GuoZXWCC20,DBLP:conf/iclr/ZhuXWHQZLL20}. Recently, pre-training standard sequence-to-sequence (Seq2Seq) models has shown significant improvements and become a popular paradigm for NMT tasks
\citep{DBLP:conf/icml/SongTQLL19,DBLP:journals/tacl/LiuGGLEGLZ20,DBLP:conf/emnlp/LinPWQFZL20}.



\begin{table}[t]
\small
\setlength{\tabcolsep}{1.6pt}
\centering
\begin{tabular}{lcccc}
\hline
Approach      & Enc.    & Dec.        & Mono. & Para. \\
\hline
mBERT \cite{DBLP:conf/naacl/DevlinCLT19}        & \exis  & \noexis         & \exis       & \noexis\\
XLM \cite{DBLP:conf/nips/ConneauL19}            & \exis  & \noexis         & \exis       & \exis\\
MASS \cite{DBLP:conf/icml/SongTQLL19}           & \exis  & $\rightarrow$   & \exis       & \noexis\\
mBART \cite{DBLP:journals/tacl/LiuGGLEGLZ20}    & \exis  & $\rightarrow$   & \exis       & \noexis \\
mRASP \cite{DBLP:conf/emnlp/LinPWQFZL20}        & \exis  & $\rightarrow$   & \noexis     & \exis \\
\hline
\ours (Ours)              & \exis      & $\Longleftrightarrow$    & \exis       & \exis      \\
\hline
\end{tabular}
\caption{Comparison and summary of existing pre-trained models for machine translation. Enc: encoder; Dec: decoder; Mono: monolingual; Para: bilingual. “\exis” \ denotes the corresponding model is pre-trained or the corresponding data is used. “$\rightarrow$” denotes the decoder of model is unidirectional, “$\Longleftrightarrow$” denotes the decoder is bidirectional.}
\label{tab:compare_method}
\end{table}

\begin{figure*}[htbp]
     \centering
     \includegraphics[width=1.0\textwidth]{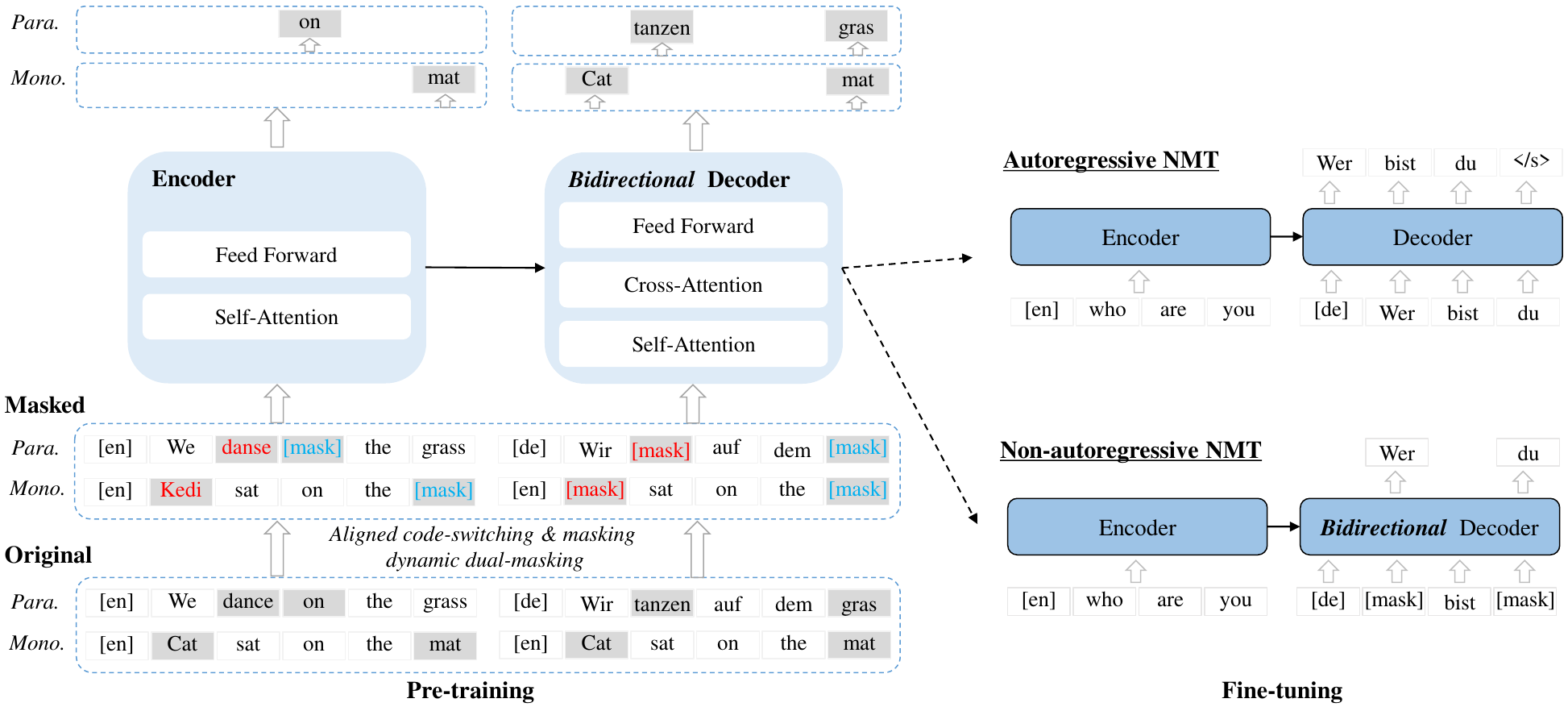}
     \caption{
    The framework for \ours, which consists of an encoder and a \textbf{\emph{bidirectional decoder}}. “\emph{Mono}” denotes monolingual, “\emph{Para}” denotes bilingual. During the pre-training (left), the original monolingual and bilingual inputs in many languages are augmented (the words are replaced with new words with same semantics or “[mask]”, please see Figure~\ref{fig:frame_work_exam} for more details) and fed into the model. Finally, we predict all the “{[mask]}” words on the source side and target side respectively. For fine-tuning (right), \ours provides unified initial parameter sets for \subANMT and \subNAT.
     }
     \label{fig:frame_work}
\end{figure*}


However, some experimental results from XLM \cite{DBLP:conf/nips/ConneauL19} have shown that the decoder module initialized by the pre-trained bidirectional masked language model (MLM) \cite{DBLP:conf/naacl/DevlinCLT19}, rather than the unidirectional causal language model (CLM, \citealp{Radford2018ImprovingLU}), would achieve better results on Autoregressive NMT (\subANMT). Especially, compared to random initialization, initialized by GPT \cite{Radford2018ImprovingLU} might result in performance degradation sometimes. We conjecture that when fine-tuning on generation tasks (e.g., NMT), the representation capability of the pre-trained models may be more needed than the generation capability. Therefore, during pre-training, we should focus on training the representation capability not only for the encoder, but also for the decoder more explicitly.


Inspired by that, we present \ours, a \textbf{multilingual Conditional masked language prE-training model for MAchine Translation}, which consists of a bidirectional encoder, a \emph{bidirectional} decoder, and a cross-attention module for bridging them. Specifically, the model is jointly trained by MLM on the encoder and Conditional MLM (CMLM) on the decoder with large-scale monolingual and bilingual texts in many languages. Table~\ref{tab:compare_method} compares our model with prior works. Benefiting from the structure, \ours can provide  unified initialization parameters not only for \subANMT task, but also for Non-autoregressive NMT (\subNAT) directly. \subNAT has been attracting more and more attention because of its feature of parallel decoding, which helps to greatly reduce the translation latency.

To better train the representation capability of the model, the masking operations are applied in two steps. First, some source words that have been aligned with target words are randomly selected and then substituted by new words of similar meanings in other languages, and their corresponding target words are masked. We call this method \aligncs. Then, the remaining words in both source and target languages will be masked by \dymask.

Extensive experiments on downstream \subANMT and \subNAT tasks show significant gains over prior works. Specifically, under low-resource conditions ($\textless$ 1M bitext pairs), our system gains up to +14.4 BLEU points over baselines. Even for extremely high-resource settings ($\textgreater$ 25M), \ours still achieves significant improvements. In addition, experiments on the WMT16 Romanian\undirec English task demonstrate that our system can be further improved (+2.1 BLEU) by the Back-Translation (BT; \citealp{DBLP:conf/acl/SennrichHB16}).

The main contributions of our work can be summarized as follows:
\begin{itemize}

    \item We propose a multilingual pre-trained model \ours, which consists of a bidirectional encoder, a \emph{bidirectional} decoder. The model is pre-trained on both monolingual and bilingual corpora and then used for initializing downstream \subANMT and \subNAT tasks. To the best of our knowledge, this is the first work to pre-train a unified model suitable for both \subANMT and \subNAT. 

    \item We introduce a two-step masking strategy to enhance the model training under the setting of bidirectional decoders. Based on a multilingual translation dictionary and word alignment between source and target sentences, \aligncs is firstly applied. Then, \dymask is used. 
	
	\item We carry out extensive experiments on \subANMT and \subNAT tasks with data of varied sizes. Consistent improvements over strong competitors demonstrate the effectiveness of \ours. 
\end{itemize}

\begin{figure*}[htbp]
    \centering
    \includegraphics[scale=0.65]{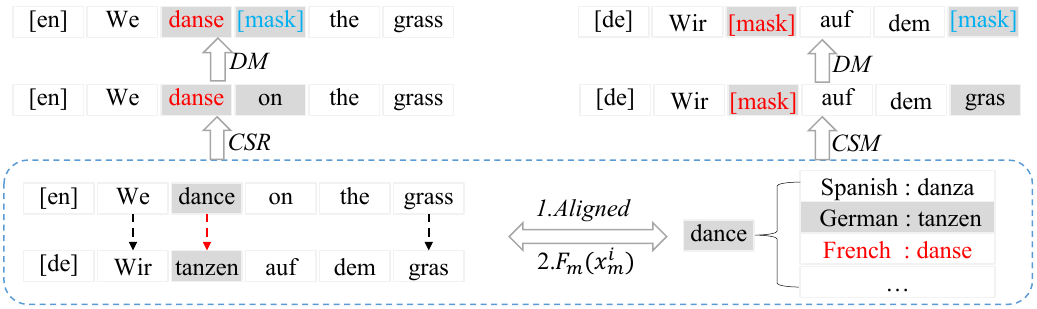}
    \caption{
    The details of our two-step masking.  
    We first obtain the aligned pair set $\Lambda=$ \{(``dance'',``tanzen''),...\} (marked with $\dashrightarrow$) from the original inputs by looking up the cross-lingual dictionary (denote as \emph{1.Aligned}), and then randomly select a subset (marked as ``dance''$\dashrightarrow$``tanzen'' with red color) from it, in the lower left of the figure. For each element in the subset, we select a new word by \emph{$F_m(x_m^i)$}, and  perform \emph{CSR} to replace the source fragment (“danse” marked as red color) and \emph{CSM} for target (“[mask]” marked as red color) respectively. Finally, we do the DM process to mask the contents of the source and target respectively (“[mask]” marked as light-blue color).
    }
    \label{fig:frame_work_exam}
\end{figure*}

\section{Pre-training Approach}

Our \ours is jointly trained by MLM and CMLM on the source side and the target side, respectively. The overall framework is illustrated in Figure~\ref{fig:frame_work}. In this section, we first introduce the multilingual CMLM task (Section~\ref{sec:multi_cmlm}). Then, we describe the two-step masking, including the \aligncs (Section~\ref{sec:aligned_cs_and_m}) and the \dymask (Section~\ref{sec:dyna_mask}). Finally, we present training objectives of \ours  (Section~\ref{sec:obj}).

Formally, our training data consists of $M$ language-pairs $D = \{D_1,D_2,...,D_M\}$. $D_k(m,n)$ is a collection of sentence pairs in language $L_m$ and $L_n$, respectively. In the description below, we denote a sentence pair as $(X_m,Y_n) \in D_k(m,n)$, where $X_m$ is the source text in the language $L_m$, and $Y_n$ is the corresponding target text in the language $L_n$. For monolingual corpora, we create pseudo bilingual text by copying the sentence, namely, $X_m=Y_n$.

\subsection{Conditional Masked Language Model}
\label{sec:multi_cmlm}

CMLM predicts masked tokens $y_n^{mask}$, given a source sentence $X_m$ and the remaining target sentence $Y_n\backslash y_n^{mask}$
. The probability of each $y_n^j \in y_n^{mask}$ is independently calculated:  
\begin{equation}
    P(y_n^j|X_m,Y_n\backslash y_n^{mask}).
\end{equation}

CMLM can be directly used to train a standard Seq2Seq model with a bidirectional encoder, a unidirectional decoder, and a cross attention. However, it is not restricted to the autoregressive feature on the decoder side because of the independence between masked words. Therefore, following practices of \subNAT, we use CMLM to pre-train a Seq2Seq model with a bidirectional decoder, as shown in Figure~\ref{fig:frame_work}.  

Although bilingual sentence pairs can be directly used to train the model together with the conventional CMLM \cite{DBLP:conf/emnlp/Ghazvininejad2019}, it is challenging for sentence pairs created from monolingual corpora because of identical source and target sentences. Therefore, we introduce a two-step masking strategy to enhance model training on both bilingual and monolingual corpora.



\subsection{Aligned Code-Switching \& Masking}
\label{sec:aligned_cs_and_m}

We use \aligncs strategy to replace the source word or phrase with a new word in another language, and then mask the corresponding target word. Different from the previous code-switching methods \cite{DBLP:conf/emnlp/yang20, DBLP:conf/emnlp/LinPWQFZL20} where source words always are randomly selected and replaced directly, our method consists of three steps: 

\begin{enumerate}
    \item \textbf{Aligning}: We utilize a multilingual translation dictionary to get a set of aligned words $\Lambda = \{\cdots,(x_m^i,y_n^j),\cdots\}$ between the source $X_m$ and target $Y_n$. The word pair $(x_m^i,y_n^j)$ denotes that the $i$-th word in $X_m$ and $j$-th word in $Y_n$ are translations of each other. For sentence pairs created from monolingual corpora, words in an aligned word pair are identical.
    
    \item \textbf{Code-Switching Replace (CSR)}: Given an aligned word pair $(x_m^i,y_n^j) \in \Lambda$, we first select a new word $\hat{x}_k^i$ in the language $L_k$ that can be used to replace $x_m^i$ in the source sentence $X_m$, 
$$\hat{x}_k^i = F_{m}(x_m^i)$$ 
where $F_{m}(x)$ is a multilingual dictionary lookup function for a word $x$ in the language $L_m$, $\hat{x}_k^i$ is a randomly selected word from the dictionary, which is a translation of $x_m^i$ in the language $L_k$. 

    \item \textbf{Code-Switching Masking (CSM)}: If the source word $x_m^i$ in the aligned pair $(x_m^i,y_n^j)$ is replaced by $\hat{x}_k^i$, we also mask $y_n^j$ in $Y_n$ by replacing it with a universal mask token. Then, \ours will be trained to predict it in the output layers of the bidirectional decoder.
\end{enumerate}

For aligning and CSR, we only use available multilingual translation dictionary provided by MUSE \cite{DBLP:conf/arxiv/Lample2018}. Figure~\ref{fig:frame_work_exam} shows the process of \aligncs. According to the given dictionary, “dance” and “tanzen” are aligned, then a new French word “danse” is selected to replace “dance”, and “tanzen” replaced by “[mask]” (marked as red color).

During training, at most 15\% of the words in the sentence will be performed by CSR and CSM. For monolingual data, we set this ratio to 30\%. We use $$(\mathrm{CSR}(X_m), \mathrm{CSM}(Y_n))$$ to denote the new sentence pair after \aligncs, which will be further dynamically dual-masked at random.

\subsection{Dynamic Dual-Masking}
\label{sec:dyna_mask}

Limited by the dictionary, the ratio of aligned word pairs is usually small. In fact, we can only match aligned pairs for 6\% of the tokens on average in the bilingual corpora. To further increase the training efficiency, we perform \dymask (DM) on both bilingual and monolingual data. 

\begin{itemize}
    \item Bilingual data: We first sample a masking ratio $\upsilon$ from a uniform distribution between $[0.2,0.5]$, then randomly select a subset of target words which are replaced by ``[mask]''. Similarly, we select a subset on the source texts and mask them with a ratio of $\mu$ in a range of $[0.1,0.2]$. Figure~\ref{fig:frame_work_exam} shows an example of \dymask on bilingual data. We set $\upsilon \geq \mu$ to force the bidirectional decoder to obtain more information from the encoder.
    
    \item Monolingual data: Since the source and target are identical before masking, we sample $\upsilon=\mu$ from a range $[0.3,0.4]$ and mask the same subset of words on both sides. This will avoid the decoder directly copying the token from the source.
\end{itemize}

Follow practices of pre-trained language models, 10\% of the selected words for masking remain unchanged, and 10\% replaced with a random token. Words replaced by the \aligncs will not be selected to prevent the loss of cross-lingual information. We use $$(\mathrm{DM}(\mathrm{CSR}(X_m)), \mathrm{DM}(\mathrm{CSM}(Y_n)))$$ to denote the new sentence pair after dynamic dual-masking, which will be used for pre-training.

\subsection{Multilingual Pre-training Objectives}
\label{sec:obj}

We jointly train the encoder and decoder on MLM and CMLM tasks. Given the sentence pair $$(\hat{X}_m,\hat{Y}_n)=(\mathrm{DM}(\mathrm{CSR}(X_m)), \mathrm{DM}(\mathrm{CSM}(Y_n)))$$ from the masked corpora $\hat{D}$, the final training objective is formulated as follows:
\begin{equation}
\begin{aligned}
\mathcal{L} = -\sum_{(\hat{X}_m,\hat{Y}_n)\in \hat{D}} \lambda &  \sum_{y_n^j\in y_n^{mask}} \log P(y_n^j|\hat{X}_m,\hat{Y}_n) \\ 
 + (1 - \lambda) & \sum_{x_m^i\in x_m^{mask}} \log P(x_m^i|\hat{X}_m)
\end{aligned}
\end{equation}
where $y_n^{mask}$ are the set of masked target words, $x_m^{mask}$ are the set of masked source words, and $\lambda$ is a hyper-parameter to balance the influence of both tasks. In our experiments, we set $\lambda=0.7$.

\begin{table*}[htbp]
\centering
\resizebox{\textwidth}{!}{
\begin{tabular}{lccccccccccccccc}
\hline
Lang-Pairs & \multicolumn{2}{c}{En-Kk} & \multicolumn{2}{c}{En-Tr} & \multicolumn{2}{c}{En-Et} & \multicolumn{2}{c}{En-Fi} & \multicolumn{2}{c}{En-Lv}    & En-Cs   & En-De     & En-Fr   & Avg \\   
Source     & \multicolumn{2}{c}{WMT19} & \multicolumn{2}{c}{WMT17} & \multicolumn{2}{c}{WMT18} & \multicolumn{2}{c}{WMT17}  & \multicolumn{2}{c}{WMT17}    & WMT19   & WMT19     & WMT14    & \\  
Size       & \multicolumn{2}{c}{91k(low)}   & \multicolumn{2}{c}{207k(low)}  & \multicolumn{2}{c}{1.94M(medium)}  & \multicolumn{2}{c}{2.66M(medium)} & \multicolumn{2}{c}{4.5M(medium)}     & 11M(high)    & 38M(extr-high)       & 41M(extr-high)     &  \\  
Direction  & $\rightarrow$ & $\leftarrow$  & $\rightarrow$ & $\leftarrow$    & $\rightarrow$ & $\leftarrow$    & $\rightarrow$ & $\leftarrow$  & $\rightarrow$ & $\leftarrow$  & $\rightarrow$  & $\rightarrow$   & $\rightarrow$ \\
\hline
Direct & 0.2  & \phantom{0}0.8    & \phantom{0}9.5   & 12.2  & 17.9  & 22.6 & 20.2  & 21.8  & 12.9 & 15.6   & 16.5  & 30.9  & 41.4    &17.1  \\
mBART  & 2.5  & \phantom{0}7.4    & 17.8  & 22.5  & 21.4  & 27.8 & 22.4  & 28.5  & 15.9 & 19.3   & 18.0   & 30.5   & 41.0  & 21.2  \\
mRASP  & 8.3  & 12.3   & 20.0  & 23.4  & 20.9  & 26.8 & 24.0  & 28.0  & 21.6 & \textbf{24.4}   & 19.9  & 35.2  & \textbf{44.3} & 23.8  \\
\ours        & \textbf{8.8}  & \textbf{12.9}   & \textbf{23.9}  & \textbf{23.6}  & \textbf{22.2}  & \textbf{28.5} & \textbf{25.4}  & \textbf{28.7}  & \textbf{22.0} & 24.3   & \textbf{21.5}  & \textbf{39.2}  & 43.7 & 25.0 \\
$\Delta$     & +8.6  & +12.1   & +14.4  & +11.4  & +4.3   & +5.9  & +5.2   & +6.9   & +9.1  & +8.7    & +5.0   & +8.3   & +2.3  & +7.9 \\
\hline
\end{tabular}}
\caption{Comprehensive comparison with mRASP and mBART. Best results are highlighted in \textbf{bold}. \ours outperforms them on \subANMT for all language pairs but two directions. Even for extremely high-resource scenarios(denoted as “extr-high”), we observe gains of up to +8.3 BLEU on En\undirec De language pair.  
}
\label{tab:compare_mrasp}
\end{table*}

\section{Pre-training Settings}

\paragraph{Pre-training Data} We use the English-centric multilingual parallel corpora of PC32$\footnote{\url{https://github.com/linzehui/mRASP}}$, and then collect 21-language monolingual corpora from common crawl$\footnote{\url{https://commoncrawl.org/}}$. In this paper, we use ISO language code$\footnote{\url{https://www.loc.gov/standards/iso639-2/php/code\_list.php}}$ to identify each language. A ``[{\it language code}]'' token will be prepended to the beginning of the source and target sentence as shown in Figure~\ref{fig:frame_work_exam}. This type of token helps the model to distinguish sentences from different languages. The detailed correspondence and summary of our pre-training corpora can be seen in Appendix~\ref{append:data}.



\paragraph{Data pre-processing} We directly learn a shared BPE  \cite{DBLP:conf/acl/SennrichHB16a} model on the entire data sets after tokenization. We apply Moses tokenization \cite{DBLP:conf/acl/SennrichHB16a} for most languages, and for other languages, we use KyTea\footnote{\url{http://www.phontron.com/kytea/}} for Japanese and jieba\footnote{\url{https://github.com/fxsjy/jieba}} for Chinese, and a special normalization for Romanian  \cite{DBLP:conf/acl/SennrichHB16}. Following \citet{DBLP:journals/tacl/LiuGGLEGLZ20}, we balance the vocabulary size of languages by up/down-sampling text based on their data size when learning BPE.

\paragraph{Model and Settings} As shown in Figure~\ref{fig:frame_work}, we apply a bidirectional decoder so that it can utilize left and right contexts to predict each token. We use a 6-layer encoder and 6-layer bidirectional decoder with a model dimension of 1024 and 16 attention heads. Following \citet{DBLP:conf/nips/VaswaniSPUJGKP17}, we use sinusoidal positional embedding, and apply layer normalization for word embedding and pre-norm residual connection following \citet{DBLP:conf/acl/WangLXZLWC19}.

Our model is trained on 32 Nvidia V100 GPUs for 300K steps, The batch size on each GPU is 4096 tokens, and we set the value of update frequency to 8. Following the training settings in Transformer, we use Adam optimizer ($\epsilon=1e-6, \beta_{1}=0.9, \beta_{2}=0.98$) and polynomial decay scheduling with a warm-up step of 10,000.

\section{\ANMT}
\label{sec:ANMT}


In this section,  we verify \ours provides consistent performance gains in low to extremely high resource scenarios. We also compare our method with other existing pre-training methods and further present analysis for better understanding the contributions of each component.

\subsection{Fine-Tuning Objective}
The \subANMT model consists of an encoder and a unidirectional decoder. The encoder maps a source sentence $X_m$ into hidden representations which are then fed into the decoder. The unidirectional decoder predicts the $t$-th token in a target language $L_n$ conditioned on $X_m$ and the previous target tokens $y_n^{\textless t}$. The training objective of \subANMT is to minimize the negative log-likelihood:
\begin{equation}
\begin{aligned}
& \mathcal{L}(\theta) =  \\
& \sum_{(X_m,Y_n) \in D(m,n)} \sum_{t=1}^{|Y_n|}-\log  P(y_n^t|X_m,y_n^{\textless t};\theta)
\label{eq:anmt_goal}
\end{aligned}
\end{equation}

\subsection{Experimental Settings}

\paragraph{Benchmarks} We selected 9 different language pairs and then use \ours to fine-tune on them. They are divided into four categories according to their data size: low-resource ($\textless$ 1M), medium-resource ($\textgreater$  1M and $\textless$ 10M), high-resource ($\textgreater$  10M and $\textless$  25M), and extremely high-resource ($\textgreater$  25M). See Appendix~\ref{append:scenarios} for more details.

\paragraph{Configuration} 
We adopt a dropout rate of 0.1 for extremely high-resource En\undirec Fr, En\undirec De (WMT19); for all other language pairs, we set the value of 0.3.  We fine-tune \subANMT with a maximum learning rate of $5e-4$, a warm-up step of 4000 and label smoothing of 0.2. For inference, we use beam search with a beam size of 5 for all translation directions. For a fair comparison with previous works, all results are reported with case-sensitive and tokenized BLEU scores.

\subsection{Results and Analysis}
\paragraph{Main Results}

We fine-tune \subANMT systems initialized by our \ours on 8 popular language pairs, which are the overlapping language pairs in experiments of mBART \cite{DBLP:journals/tacl/LiuGGLEGLZ20} and mRASP \cite{DBLP:conf/emnlp/LinPWQFZL20}. Table~\ref{tab:compare_mrasp} shows the results. Compared to directly training \subANMT models, our systems with \ours as initialization obtain significant improvements on all four scenarios. We observe gains of up to +14.4 BLEU and over +11.4 BLEU on three of the four tasks on low-resource scenarios, i.e., En\bidirec Tr. Without loss of generality, as the scale of the dataset increases, the benefits of pre-training models are getting smaller and smaller. However, we can still obtain significant gains when the data size is large enough (extremely high-resource: $\textgreater$ 25M), i.e. +8.3 and +2.3 BLEU for En\undirec De and En\undirec Fr respectively. This notable improvement shows that our model can further enhance extremely high-resource translation. Overall, we obtain performance gains of more than +8.0 BLEU for most directions, and finally observe gains of +7.9 BLEU on average on all language pairs. 

We further compare our \ours with mBART \cite{DBLP:journals/tacl/LiuGGLEGLZ20} and mRASP \cite{DBLP:conf/emnlp/LinPWQFZL20}, which are two pre-training methods of current SOTA. As illustrated in Table~\ref{tab:compare_mrasp}, \ours outperforms mBART on all language pairs with a large margin (+3.8 BLEU on average), for extremely high-resource, we can obtain significant improvements when mBART hurts the performance. Compared to mRASP, we achieve better performance on 11 out of the total 13 translation directions, and outperforms this strong competitor with an average improvement of +1.2 BLEU on all directions.

\paragraph{Comparison with Existing Pre-training Models}

We further compare our \ours with more existing multilingual pre-trained models on three popular translation directions, including WMT14 En\undirec De, WMT16 En\bidirec Ro. Results are shown in Table~\ref{tab:comapre_others}. Our \ours obtains competitive results on these languages pairs on average, and achieves the best performance on En\undirec Ro. 

Our model also outperforms BT \cite{DBLP:conf/acl/SennrichHB16}, which is a universal and stable approach to augment bilingual with monolingual data. In addition, when combining back-translation with our \ours on Ro \undirec En, we obtain a significantly improvement from 36.8 to 39.0 BLEU, as shown in Table~\ref{tab:comapre_others}. This indicates that our method is complementary to BT.


\paragraph{The Effectiveness of Aligned Code-Switching and Masking} We investigate the effectiveness of \aligncs as shown in Table~\ref{table:analysis}. We find that utilizing \aligncs can help \ours improve the performance for all different scenarios with gains of +0.5 BLEU on average, even though we can only match the aligned word pairs for 6\% of the tokens on average in the bilingual corpora. We presume the method can be improved more significantly if we adopt more sophisticated word alignment methods.

\paragraph{The Effectiveness of Dynamic Masking} In the pre-training phase, we use a dynamic strategy when doing dual-masking on the encoder and decoder respectively. We verify the effectiveness of this dynamic masking strategy. As illustrated in Table~\ref{table:analysis} and Appendix~\ref{append:ablations}, we achieve significant gains with margins from +0.4 to +4.5 BLEU, when we adjusted the ratio of masking from a static value to a dynamically and randomly selected value. The average improvement on all language pairs is +2.1 BLEU. This suggests the importance of dynamic masking.

\begin{table}[htbp]
\setlength\tabcolsep{4pt}
\small
\centering
\begin{tabular}{lccccccc}
\hline
Lang-Pairs  & En $\rightarrow$ De & En $\rightarrow$ Ro & Ro $\rightarrow$ En & Ro $\rightarrow$ En\\
Size        & 4.5M            &  597K             & 597K              & ( +BT ) \\
\hline    
Direct                                          & 29.3              & 34.3              & 34.0                & 36.8 \\
mBART & -                 & 37.7              & 37.8              & 38.8  \\
mRASP & 30.3              & 37.6              & 36.9              & 38.9  \\
MASS & 28.9              & --                & --                & 39.1 \\
XLM & 28.8              & --                & 35.6              & 38.5 \\
mBERT & 28.6              & --                & --                & --   \\
\hline     
\ours             & 30.0                        & 38.0              & 37.1             & 39.0  \\
\hline
\end{tabular}
\caption{Comparison with recent multilingual pre-training models on WMT14 En\undirec De, WMT16 En\bidirec Ro. We reach comparable results on all three directions. When combining back-translation, we further obtain gains of +2.2 BLEU on Ro\undirec En. }
\label{tab:comapre_others}
\end{table}

\begin{table*}[htbp]
\centering
\resizebox{\textwidth}{!}{
\begin{tabular}{lccccccccccc}
\hline
Lang-Pairs & \multicolumn{2}{c}{En-Kk} & \multicolumn{2}{c}{En-Tr} & \multicolumn{2}{c}{En-Et} & \multicolumn{2}{c}{En-Fi} & \multicolumn{2}{c}{En-Lv} & Avg \\
Direction  & $\rightarrow$  & $\leftarrow$ & $\rightarrow$  & $\leftarrow$ & $\rightarrow$  & $\leftarrow$ & $\rightarrow$  & $\leftarrow$ & $\rightarrow$  & $\leftarrow$ & \\
\hline
\ours     & 8.8  & 12.9   & 23.9  & 23.6   & 22.2   &  28.5  & 25.4   & 28.7    & 22.0  & 24.3  & 22.0 \\
. w/o \emph{Aligned CS masking}  & 8.0  & 12.3   & 23.6    & 23.1   & 22.1  & 28.0   & 24.8  & 28.1   & 21.4  & 24.1 & 21.5 \\
. w/o \emph{Aligned CS masking} \& \emph{Dynamic} & 7.2 & \phantom{0}8.7    & 21.2  & 20.4     & 20.8   & 26.8  & 24.4   & 27.5  & 16.9  & 20.2  & 19.4\\
\hline
\end{tabular}}
\caption{
Verification of the effectiveness of different techniques. “. w/o \emph{Aligned CS masking}” denotes that we pre-train \ours without \aligncs algorithm. “. w/o \emph{Aligned CS masking} \& \emph{Dynamic}” means that we further abandon the use of dynamic setting for dual-masking, where we only use a fixed masking ratio with 0.15 for the encoder and decoder. More details can be found in Appendix~\ref{append:ablations}. We can see two methods are all critical components.
}
\label{table:analysis}
\end{table*}

\section{Non-autoregressive Neural Machine Translation}
In this section, we will verify the performance of our \ours on the \subNAT, which generates translations in parallel, on widely-used translation tasks.

\subsection{Fine-Tuning Objective}

As illustrated in Figure~\ref{fig:frame_work},  \subNAT also adopts a Seq2Seq framework, but consists of an encoder and a bidirectional decoder which can be used to predict the target sequences in parallel. The training objective of \subNAT is formulated as follows:
\begin{equation}
\mathcal{L}(\theta) = \sum_{(X_m,Y_n) \in D(m,n)} \sum_{t=1}^{|Y_n|}-\log  P(y_n^t|X_m;\theta)
\end{equation}

In this work, we follow \citet{DBLP:conf/emnlp/Ghazvininejad2019}, which randomly sample some tokens $y_n^{mask}$ for masking from target sentences and train the model by predicting them given source sentences and remaining targets. The training objective is:
\begin{equation}
\begin{aligned}
& \mathcal{L}(\theta) = \sum_{(X_m,Y_n) \in D(m,n)} \\
&  \sum_{y_n^j\in y_n^{mask}}-\log P(y_n^j|X_m,Y_n\backslash y_n^{mask};\theta)
\end{aligned}
\end{equation}

During decoding, given an input sequence to translate, the initial decoder input is a sequence of ``[mask]'' tokens. The fine-tuned model generates translations by iteratively predicting target tokens and masking low-quality predictions. This process can make the model re-predict the more challenging cases conditioned on previous high-confidence predictions.

\subsection{Experimental Settings}

\paragraph{\subNAT Benchmark Data} We evaluate on three popular datasets: WMT14 En\bidirec De, WMT16 En\bidirec Ro and IWSLT14 En\bidirec De. For a fair comparison with baselines, we only use the bilingual PC32 corpora to pre-train our \ours. We only use knowledge distillation \cite{DBLP:conf/iclr/Gu0XLS18} on WMT14 En\bidirec De tasks.

\paragraph{Baselines} We use our \ours for initialization and fine-tune a Mask-Predict model \cite{DBLP:conf/emnlp/Ghazvininejad2019} as in Section~\ref{sec:ANMT}. To better quantify the effects of the proposed pre-training models, we build two strong baselines.

\textbf{Direct.} We directly train a Mask-Predict model with randomly initialized parameters.

\textbf{mRASP.} To verify that our pre-trained model is more suitable for \subNAT, we use a recently pre-trained model mRASP \cite{DBLP:conf/emnlp/LinPWQFZL20} to fine-tune on downstream language pairs.

\begin{table*}[htbp]
\centering
\resizebox{\textwidth}{!}{
\begin{tabular}{l|ccccccc}
\hline
Source  & \multicolumn{2}{c}{IWSLT14}   & \multicolumn{2}{c}{WMT16}     & \multicolumn{2}{c}{WMT14}  & Avg  \\
Lang-Pairs        & En$\rightarrow$De         & De$\rightarrow$En         & En$\rightarrow$Ro         & Ro$\rightarrow$En         & En$\rightarrow$De         & De$\rightarrow$En   &      \\
\hline
Transformer \cite{DBLP:conf/nips/VaswaniSPUJGKP17} & 23.9          & 32.8          & 34.1          & 34.5          & 28.0          & 32.7       &  31.0 \\
\hline
Mask-Predict \cite{DBLP:conf/emnlp/Ghazvininejad2019}  & 22.0          & 28.4          & 31.5          & 31.7          & 26.1          & 29.0     & 28.1     \\
mRASP \cite{DBLP:conf/emnlp/LinPWQFZL20}              & 23.9          & 30.3          & 32.2          & 32.1          & 26.7          & 29.8      & 29.2  \\

\ours(Ours)                                         & \textbf{26.7} & \textbf{33.7} & \textbf{33.3} & \textbf{33.0} & \textbf{27.2}          & \textbf{29.9}      & 30.6   \\
\hline
\end{tabular}}
\caption{Comprehensive comparison with two strong baselines. “mRASP” denotes using mRASP to initialize Mask-Predict, “\ours(Ours)” denotes using our \ours to initialize. We obtain consistent and significant improvements on all language pairs, outperforming \subANMT on IWSLT14 tasks. Best non-autoregressive results are highlighted in \textbf{bold}.}
\label{tab:nat}
\end{table*}

\paragraph{Configuration} We use almost the same configuration as the pre-training and \subANMT except the following differences. We use learned positional embeddings \cite{DBLP:conf/emnlp/Ghazvininejad2019} and set the max-positions to 10,000. 

\subsection{Main Results}
The main results on three language pairs are presented in Table~\ref{tab:nat}.
When using \ours to initialize the Mask-Predict model, we observe significant improvements (from +0.9 to +5.3 BLEU) on all different tasks, and finally obtain gains of +2.5 BLEU on average. We also achieve higher results than the \subANMT model on both En\undirec De (+2.8 BLEU) and De\undirec En (+0.9 BLEU) directions on IWSLT14 datasets, which is the extremely low-resource scenarios where training from scratch is harder and pre-training is more effective.

As illustrated in Table~\ref{tab:nat}, on all different tasks, \ours outperforms mRASP with a significant margin. On average, we obtain gains of +1.4 BLEU over mRASP. Especially under low-resource settings on IWSLT14 De$\rightarrow$En, we achieve a large gains of +3.4 BLEU over mRASP. Overall, mRASP shows limited improvement (+0.4 to +1.9 BLEU) compared to \ours. This also suggests that although we can use the traditional pre-training method to fine-tune the NAT task, it does not bring a significant improvement like the \subANMT task because of the gap between pre-training and fine-tuning tasks.

We further compare the dynamic performance on three language pairs during iterative decoding, as shown in Appendix~\ref{append:iter_NAT}. We only need 3 to 6 iterations to achieve the best score. During the iteration, we always maintain rapid improvements. In contrast, mRASP obtains the best result after 6 to 9 iterations. We also observe a phenomenon that the performance during iterations is also unstable on both mRASP and Mask-Predict, but \ours appears more stable. We conjecture that our pre-trained model can learn more related information between words in both the same and different languages. This ability alleviated the drawback of NAT assumptions: the individual token predictions are conditionally independent of each other.

\section{Related Work}

\paragraph{Multilingual Pre-training Task} 
\citet{DBLP:conf/nips/ConneauL19} and \citet{DBLP:conf/naacl/DevlinCLT19} proposed to pre-train a cross-lingual language model on multi language corpora, then the encoder or decoder of model are initialized independently for fine-tuning. \citet{DBLP:conf/icml/SongTQLL19}, \citet{DBLP:conf/emnlp/yang20} and \citet{DBLP:conf/acl/LewisLGGMLSZ20} directly pre-trained a Seq2Seq model by reconstructing part or all of inputs and achieve significant performance gains. Recently, mRASP \cite{DBLP:conf/emnlp/LinPWQFZL20} and CSP \cite{DBLP:conf/emnlp/yang20} apply the code-switching technology to simply perform random substitution on the source side. Another similar work, DICT-MLM \cite{DBLP:journals/corr/abs-2010-12566} introduce multilingual dictionary, pre-training the MLM by mask the words and then predict its cross-lingual synonyms. mRASP2 \cite{DBLP:conf/acl/PanWWL20} also used code-switching on monolingual and bilingual data to improve the effectiveness, but it is essentially a multilingual \subANMT model. 

Compared to previous works: 1) \ours is the first pre-trained Seq2Seq model with a bidirectional decoder; 2) We introduce aligned code-switching \& masking, different from traditional code-switching, we have two additional steps: align between source and target, and CSM; 3) We also introduce a dynamic dual-masking method.


\paragraph{\ANMT} Our work is also related to \subANMT, which adopts an encoder-decoder framework to train the model \cite{DBLP:conf/nips/SutskeverVL14}. To improve the performance, back-translation, forward-translation and related techniques were proposed to utilize the monolingual corpora \cite{DBLP:conf/acl/SennrichHB16,DBLP:conf/emnlp/ZhangZ16,DBLP:conf/emnlp/EdunovOAG18,DBLP:conf/aclnmt/HoangKHC18}.  
Prior works also attempted to jointly train a single multilingual translation model that translates multi-language directions at the same time \citep{DBLP:journals/corr/FiratCB16,DBLP:journals/tacl/JohnsonSLKWCTVW17,DBLP:conf/naacl/AharoniJF19, DBLP:conf/emnlp/WuLZLHL21}. 
In this work, we focus on pre-training a multilingual language model, which can provide initialization parameters for the language pairs. On the other hand, our method can use other languages to further improve high-resource tasks.


\paragraph{\NAT} \citet{DBLP:conf/iclr/Gu0XLS18} first introduced a transformer-based method to predict the complete target sequence in parallel. In order to reduce the gap with the AT model, \citet{DBLP:conf/emnlp/LeeMC18} and \citet{DBLP:conf/emnlp/Ghazvininejad2019} proposed to decode the target sentence with iterative refinement. \citet{DBLP:conf/aaai/WangTHQZL19} and \citet{DBLP:conf/nips/SunLWHLD19} utilized auxiliary information to enhance the performance of NAT. One work related to us is \citet{DBLP:conf/nips/GuoZXWCC20}, which using BERT to initialize the \subNAT. In this work, \ours is the first attempt to pre-train a multilingual Seq2Seq language model on NAT task.

\section{Conclusion}
In this paper, we demonstrate that multilingually pre-training a sequence-to-sequence model but with a bidirectional decoder produces significant performance gains for both \ANAT. Benefiting from conditional masking, the decoder module, especially the cross-attention can learn the word representation and cross-lingual representation ability more easily. We further introduce the \aligncs to align the representation space for words with similar semantics but in different languages, then we use a \dymask strategy to induce the bidirectional decoder to actively obtain the information from the source side.  Finally, we verified the effectiveness of these two methods. In the future, we will investigate more effective word alignment method for \aligncs.

\section{Acknowledgments}
We would like to thank anonymous reviewers for their helpful feedback. we also thank Wenyong Huang, Lu Hou, Yinpeng Guo, Guchun Zhang for their useful suggestion and help with experiments.

\bibliography{anthology,custom}

\begin{thebibliography}{30}
\expandafter\ifx\csname natexlab\endcsname\relax\def\natexlab#1{#1}\fi

\bibitem[{Aharoni et~al.(2019)Aharoni, Johnson, and
  Firat}]{DBLP:conf/naacl/AharoniJF19}
Roee Aharoni, Melvin Johnson, and Orhan Firat. 2019.
\newblock \href {https://doi.org/10.18653/v1/n19-1388} {Massively multilingual
  neural machine translation}.
\newblock In \emph{Proceedings of the 2019 Conference of the North American
  Chapter of the Association for Computational Linguistics: Human Language
  Technologies, {NAACL-HLT} 2019, Minneapolis, MN, USA, June 2-7, 2019, Volume
  1 (Long and Short Papers)}, pages 3874--3884. Association for Computational
  Linguistics.

\bibitem[{Chaudhary et~al.(2020)Chaudhary, Raman, Srinivasan, and
  Chen}]{DBLP:journals/corr/abs-2010-12566}
Aditi Chaudhary, Karthik Raman, Krishna Srinivasan, and Jiecao Chen. 2020.
\newblock \href {http://arxiv.org/abs/2010.12566} {{DICT-MLM:} improved
  multilingual pre-training using bilingual dictionaries}.
\newblock \emph{CoRR}, abs/2010.12566.

\bibitem[{Conneau and Lample(2019)}]{DBLP:conf/nips/ConneauL19}
Alexis Conneau and Guillaume Lample. 2019.
\newblock \href
  {https://proceedings.neurips.cc/paper/2019/hash/c04c19c2c2474dbf5f7ac4372c5b9af1-Abstract.html}
  {Cross-lingual language model pretraining}.
\newblock In \emph{Advances in Neural Information Processing Systems 32: Annual
  Conference on Neural Information Processing Systems 2019, NeurIPS 2019,
  December 8-14, 2019, Vancouver, BC, Canada}, pages 7057--7067.

\bibitem[{Devlin et~al.(2019)Devlin, Chang, Lee, and
  Toutanova}]{DBLP:conf/naacl/DevlinCLT19}
Jacob Devlin, Ming{-}Wei Chang, Kenton Lee, and Kristina Toutanova. 2019.
\newblock \href {https://doi.org/10.18653/v1/n19-1423} {{BERT:} pre-training of
  deep bidirectional transformers for language understanding}.
\newblock In \emph{Proceedings of the 2019 Conference of the North American
  Chapter of the Association for Computational Linguistics: Human Language
  Technologies, {NAACL-HLT} 2019, Minneapolis, MN, USA, June 2-7, 2019, Volume
  1 (Long and Short Papers)}, pages 4171--4186. Association for Computational
  Linguistics.

\bibitem[{Edunov et~al.(2018)Edunov, Ott, Auli, and
  Grangier}]{DBLP:conf/emnlp/EdunovOAG18}
Sergey Edunov, Myle Ott, Michael Auli, and David Grangier. 2018.
\newblock \href {https://doi.org/10.18653/v1/d18-1045} {Understanding
  back-translation at scale}.
\newblock In \emph{Proceedings of the 2018 Conference on Empirical Methods in
  Natural Language Processing, Brussels, Belgium, October 31 - November 4,
  2018}, pages 489--500. Association for Computational Linguistics.

\bibitem[{Firat et~al.(2016)Firat, Cho, and
  Bengio}]{DBLP:journals/corr/FiratCB16}
Orhan Firat, KyungHyun Cho, and Yoshua Bengio. 2016.
\newblock \href {http://arxiv.org/abs/1601.01073} {Multi-way, multilingual
  neural machine translation with a shared attention mechanism}.
\newblock \emph{CoRR}, abs/1601.01073.

\bibitem[{Ghazvininejad et~al.(2019)Ghazvininejad, Levy, Liu, and
  Zettlemoyer}]{DBLP:conf/emnlp/Ghazvininejad2019}
Marjan Ghazvininejad, Omer Levy, Yinhan Liu, and Luke Zettlemoyer. 2019.
\newblock \href {https://doi.org/10.18653/v1/D19-1633} {Mask-predict: Parallel
  decoding of conditional masked language models}.
\newblock In \emph{Proceedings of the 2019 Conference on Empirical Methods in
  Natural Language Processing and the 9th International Joint Conference on
  Natural Language Processing, {EMNLP-IJCNLP} 2019, Hong Kong, China, November
  3-7, 2019}, pages 6111--6120. Association for Computational Linguistics.

\bibitem[{Gu et~al.(2018)Gu, Bradbury, Xiong, Li, and
  Socher}]{DBLP:conf/iclr/Gu0XLS18}
Jiatao Gu, James Bradbury, Caiming Xiong, Victor O.~K. Li, and Richard Socher.
  2018.
\newblock \href {https://openreview.net/forum?id=B1l8BtlCb} {Non-autoregressive
  neural machine translation}.
\newblock In \emph{6th International Conference on Learning Representations,
  {ICLR} 2018, Vancouver, BC, Canada, April 30 - May 3, 2018, Conference Track
  Proceedings}. OpenReview.net.

\bibitem[{Guo et~al.(2020)Guo, Zhang, Xu, Wei, Chen, and
  Chen}]{DBLP:conf/nips/GuoZXWCC20}
Junliang Guo, Zhirui Zhang, Linli Xu, Hao{-}Ran Wei, Boxing Chen, and Enhong
  Chen. 2020.
\newblock \href
  {https://proceedings.neurips.cc/paper/2020/hash/7a6a74cbe87bc60030a4bd041dd47b78-Abstract.html}
  {Incorporating {BERT} into parallel sequence decoding with adapters}.
\newblock In \emph{Advances in Neural Information Processing Systems 33: Annual
  Conference on Neural Information Processing Systems 2020, NeurIPS 2020,
  December 6-12, 2020, virtual}.

\bibitem[{Hoang et~al.(2018)Hoang, Koehn, Haffari, and
  Cohn}]{DBLP:conf/aclnmt/HoangKHC18}
Cong Duy~Vu Hoang, Philipp Koehn, Gholamreza Haffari, and Trevor Cohn. 2018.
\newblock \href {https://doi.org/10.18653/v1/w18-2703} {Iterative
  back-translation for neural machine translation}.
\newblock In \emph{Proceedings of the 2nd Workshop on Neural Machine
  Translation and Generation, NMT@ACL 2018, Melbourne, Australia, July 20,
  2018}, pages 18--24. Association for Computational Linguistics.

\bibitem[{Johnson et~al.(2017)Johnson, Schuster, Le, Krikun, Wu, Chen, Thorat,
  Vi{\'{e}}gas, Wattenberg, Corrado, Hughes, and
  Dean}]{DBLP:journals/tacl/JohnsonSLKWCTVW17}
Melvin Johnson, Mike Schuster, Quoc~V. Le, Maxim Krikun, Yonghui Wu, Zhifeng
  Chen, Nikhil Thorat, Fernanda~B. Vi{\'{e}}gas, Martin Wattenberg, Greg
  Corrado, Macduff Hughes, and Jeffrey Dean. 2017.
\newblock \href {https://transacl.org/ojs/index.php/tacl/article/view/1081}
  {Google's multilingual neural machine translation system: Enabling zero-shot
  translation}.
\newblock \emph{Trans. Assoc. Comput. Linguistics}, 5:339--351.

\bibitem[{Lample et~al.(2018)Lample, Conneau, Denoyer, and
  Ranzato}]{DBLP:conf/arxiv/Lample2018}
Guillaume Lample, Alexis Conneau, Ludovic Denoyer, and Marc'Aurelio Ranzato.
  2018.
\newblock \href {https://openreview.net/forum?id=rkYTTf-AZ} {Unsupervised
  machine translation using monolingual corpora only}.
\newblock In \emph{6th International Conference on Learning Representations,
  {ICLR} 2018, Vancouver, BC, Canada, April 30 - May 3, 2018, Conference Track
  Proceedings}. OpenReview.net.

\bibitem[{Lee et~al.(2018)Lee, Mansimov, and Cho}]{DBLP:conf/emnlp/LeeMC18}
Jason Lee, Elman Mansimov, and Kyunghyun Cho. 2018.
\newblock \href {https://doi.org/10.18653/v1/d18-1149} {Deterministic
  non-autoregressive neural sequence modeling by iterative refinement}.
\newblock In \emph{Proceedings of the 2018 Conference on Empirical Methods in
  Natural Language Processing, Brussels, Belgium, October 31 - November 4,
  2018}, pages 1173--1182. Association for Computational Linguistics.

\bibitem[{Lewis et~al.(2020)Lewis, Liu, Goyal, Ghazvininejad, Mohamed, Levy,
  Stoyanov, and Zettlemoyer}]{DBLP:conf/acl/LewisLGGMLSZ20}
Mike Lewis, Yinhan Liu, Naman Goyal, Marjan Ghazvininejad, Abdelrahman Mohamed,
  Omer Levy, Veselin Stoyanov, and Luke Zettlemoyer. 2020.
\newblock \href {https://doi.org/10.18653/v1/2020.acl-main.703} {{BART:}
  denoising sequence-to-sequence pre-training for natural language generation,
  translation, and comprehension}.
\newblock In \emph{Proceedings of the 58th Annual Meeting of the Association
  for Computational Linguistics, {ACL} 2020, Online, July 5-10, 2020}, pages
  7871--7880. Association for Computational Linguistics.

\bibitem[{Lin et~al.(2020)Lin, Pan, Wang, Qiu, Feng, Zhou, and
  Li}]{DBLP:conf/emnlp/LinPWQFZL20}
Zehui Lin, Xiao Pan, Mingxuan Wang, Xipeng Qiu, Jiangtao Feng, Hao Zhou, and
  Lei Li. 2020.
\newblock \href {https://doi.org/10.18653/v1/2020.emnlp-main.210} {Pre-training
  multilingual neural machine translation by leveraging alignment information}.
\newblock In \emph{Proceedings of the 2020 Conference on Empirical Methods in
  Natural Language Processing, {EMNLP} 2020, Online, November 16-20, 2020},
  pages 2649--2663. Association for Computational Linguistics.

\bibitem[{Liu et~al.(2020)Liu, Gu, Goyal, Li, Edunov, Ghazvininejad, Lewis, and
  Zettlemoyer}]{DBLP:journals/tacl/LiuGGLEGLZ20}
Yinhan Liu, Jiatao Gu, Naman Goyal, Xian Li, Sergey Edunov, Marjan
  Ghazvininejad, Mike Lewis, and Luke Zettlemoyer. 2020.
\newblock \href {https://transacl.org/ojs/index.php/tacl/article/view/2107}
  {Multilingual denoising pre-training for neural machine translation}.
\newblock \emph{Trans. Assoc. Comput. Linguistics}, 8:726--742.

\bibitem[{Pan et~al.(2021)Pan, Wang, Wu, and Li}]{DBLP:conf/acl/PanWWL20}
Xiao Pan, Mingxuan Wang, Liwei Wu, and Lei Li. 2021.
\newblock \href {https://doi.org/10.18653/v1/2021.acl-long.21} {Contrastive
  learning for many-to-many multilingual neural machine translation}.
\newblock In \emph{Proceedings of the 59th Annual Meeting of the Association
  for Computational Linguistics and the 11th International Joint Conference on
  Natural Language Processing, {ACL/IJCNLP} 2021, (Volume 1: Long Papers),
  Virtual Event, August 1-6, 2021}, pages 244--258. Association for
  Computational Linguistics.

\bibitem[{Radford and Narasimhan(2018)}]{Radford2018ImprovingLU}
Alec Radford and Karthik Narasimhan. 2018.
\newblock Improving language understanding by generative pre-training.

\bibitem[{Sennrich et~al.(2016{\natexlab{a}})Sennrich, Haddow, and
  Birch}]{DBLP:conf/acl/SennrichHB16}
Rico Sennrich, Barry Haddow, and Alexandra Birch. 2016{\natexlab{a}}.
\newblock \href {https://doi.org/10.18653/v1/p16-1009} {Improving neural
  machine translation models with monolingual data}.
\newblock In \emph{Proceedings of the 54th Annual Meeting of the Association
  for Computational Linguistics, {ACL} 2016, August 7-12, 2016, Berlin,
  Germany, Volume 1: Long Papers}. The Association for Computer Linguistics.

\bibitem[{Sennrich et~al.(2016{\natexlab{b}})Sennrich, Haddow, and
  Birch}]{DBLP:conf/acl/SennrichHB16a}
Rico Sennrich, Barry Haddow, and Alexandra Birch. 2016{\natexlab{b}}.
\newblock \href {https://doi.org/10.18653/v1/p16-1162} {Neural machine
  translation of rare words with subword units}.
\newblock In \emph{Proceedings of the 54th Annual Meeting of the Association
  for Computational Linguistics, {ACL} 2016, August 7-12, 2016, Berlin,
  Germany, Volume 1: Long Papers}. The Association for Computer Linguistics.

\bibitem[{Song et~al.(2019)Song, Tan, Qin, Lu, and
  Liu}]{DBLP:conf/icml/SongTQLL19}
Kaitao Song, Xu~Tan, Tao Qin, Jianfeng Lu, and Tie{-}Yan Liu. 2019.
\newblock \href {http://proceedings.mlr.press/v97/song19d.html} {{MASS:} masked
  sequence to sequence pre-training for language generation}.
\newblock In \emph{Proceedings of the 36th International Conference on Machine
  Learning, {ICML} 2019, 9-15 June 2019, Long Beach, California, {USA}},
  volume~97 of \emph{Proceedings of Machine Learning Research}, pages
  5926--5936. {PMLR}.

\bibitem[{Sun et~al.(2019)Sun, Li, Wang, He, Lin, and
  Deng}]{DBLP:conf/nips/SunLWHLD19}
Zhiqing Sun, Zhuohan Li, Haoqing Wang, Di~He, Zi~Lin, and Zhi{-}Hong Deng.
  2019.
\newblock \href
  {https://proceedings.neurips.cc/paper/2019/hash/74563ba21a90da13dacf2a73e3ddefa7-Abstract.html}
  {Fast structured decoding for sequence models}.
\newblock In \emph{Advances in Neural Information Processing Systems 32: Annual
  Conference on Neural Information Processing Systems 2019, NeurIPS 2019,
  December 8-14, 2019, Vancouver, BC, Canada}, pages 3011--3020.

\bibitem[{Sutskever et~al.(2014)Sutskever, Vinyals, and
  Le}]{DBLP:conf/nips/SutskeverVL14}
Ilya Sutskever, Oriol Vinyals, and Quoc~V. Le. 2014.
\newblock \href
  {https://proceedings.neurips.cc/paper/2014/hash/a14ac55a4f27472c5d894ec1c3c743d2-Abstract.html}
  {Sequence to sequence learning with neural networks}.
\newblock In \emph{Advances in Neural Information Processing Systems 27: Annual
  Conference on Neural Information Processing Systems 2014, December 8-13 2014,
  Montreal, Quebec, Canada}, pages 3104--3112.

\bibitem[{Vaswani et~al.(2017)Vaswani, Shazeer, Parmar, Uszkoreit, Jones,
  Gomez, Kaiser, and Polosukhin}]{DBLP:conf/nips/VaswaniSPUJGKP17}
Ashish Vaswani, Noam Shazeer, Niki Parmar, Jakob Uszkoreit, Llion Jones,
  Aidan~N. Gomez, Lukasz Kaiser, and Illia Polosukhin. 2017.
\newblock \href
  {https://proceedings.neurips.cc/paper/2017/hash/3f5ee243547dee91fbd053c1c4a845aa-Abstract.html}
  {Attention is all you need}.
\newblock In \emph{Advances in Neural Information Processing Systems 30: Annual
  Conference on Neural Information Processing Systems 2017, December 4-9, 2017,
  Long Beach, CA, {USA}}, pages 5998--6008.

\bibitem[{Wang et~al.(2019{\natexlab{a}})Wang, Li, Xiao, Zhu, Li, Wong, and
  Chao}]{DBLP:conf/acl/WangLXZLWC19}
Qiang Wang, Bei Li, Tong Xiao, Jingbo Zhu, Changliang Li, Derek~F. Wong, and
  Lidia~S. Chao. 2019{\natexlab{a}}.
\newblock \href {https://doi.org/10.18653/v1/p19-1176} {Learning deep
  transformer models for machine translation}.
\newblock In \emph{Proceedings of the 57th Conference of the Association for
  Computational Linguistics, {ACL} 2019, Florence, Italy, July 28- August 2,
  2019, Volume 1: Long Papers}, pages 1810--1822. Association for Computational
  Linguistics.

\bibitem[{Wang et~al.(2019{\natexlab{b}})Wang, Tian, He, Qin, Zhai, and
  Liu}]{DBLP:conf/aaai/WangTHQZL19}
Yiren Wang, Fei Tian, Di~He, Tao Qin, ChengXiang Zhai, and Tie{-}Yan Liu.
  2019{\natexlab{b}}.
\newblock \href {https://doi.org/10.1609/aaai.v33i01.33015377}
  {Non-autoregressive machine translation with auxiliary regularization}.
\newblock In \emph{The Thirty-Third {AAAI} Conference on Artificial
  Intelligence, {AAAI} 2019, The Thirty-First Innovative Applications of
  Artificial Intelligence Conference, {IAAI} 2019, The Ninth {AAAI} Symposium
  on Educational Advances in Artificial Intelligence, {EAAI} 2019, Honolulu,
  Hawaii, USA, January 27 - February 1, 2019}, pages 5377--5384. {AAAI} Press.

\bibitem[{Wu et~al.(2021)Wu, Li, Zhang, Li, Haffari, and
  Liu}]{DBLP:conf/emnlp/WuLZLHL21}
Minghao Wu, Yitong Li, Meng Zhang, Liangyou Li, Gholamreza Haffari, and Qun
  Liu. 2021.
\newblock \href {https://doi.org/10.18653/v1/2021.emnlp-main.580}
  {Uncertainty-aware balancing for multilingual and multi-domain neural machine
  translation training}.
\newblock In \emph{Proceedings of the 2021 Conference on Empirical Methods in
  Natural Language Processing, {EMNLP} 2021, Virtual Event / Punta Cana,
  Dominican Republic, 7-11 November, 2021}, pages 7291--7305. Association for
  Computational Linguistics.

\bibitem[{Yang et~al.(2020)Yang, Hu, Han, Huang, and
  Ju}]{DBLP:conf/emnlp/yang20}
Zhen Yang, Bojie Hu, Ambyera Han, Shen Huang, and Qi~Ju. 2020.
\newblock \href {https://doi.org/10.18653/v1/2020.emnlp-main.208} {{CSP:}
  code-switching pre-training for neural machine translation}.
\newblock In \emph{Proceedings of the 2020 Conference on Empirical Methods in
  Natural Language Processing, {EMNLP} 2020, Online, November 16-20, 2020},
  pages 2624--2636. Association for Computational Linguistics.

\bibitem[{Zhang and Zong(2016)}]{DBLP:conf/emnlp/ZhangZ16}
Jiajun Zhang and Chengqing Zong. 2016.
\newblock \href {https://doi.org/10.18653/v1/d16-1160} {Exploiting source-side
  monolingual data in neural machine translation}.
\newblock In \emph{Proceedings of the 2016 Conference on Empirical Methods in
  Natural Language Processing, {EMNLP} 2016, Austin, Texas, USA, November 1-4,
  2016}, pages 1535--1545. The Association for Computational Linguistics.

\bibitem[{Zhu et~al.(2020)Zhu, Xia, Wu, He, Qin, Zhou, Li, and
  Liu}]{DBLP:conf/iclr/ZhuXWHQZLL20}
Jinhua Zhu, Yingce Xia, Lijun Wu, Di~He, Tao Qin, Wengang Zhou, Houqiang Li,
  and Tie{-}Yan Liu. 2020.
\newblock \href {https://openreview.net/forum?id=Hyl7ygStwB} {Incorporating
  {BERT} into neural machine translation}.
\newblock In \emph{8th International Conference on Learning Representations,
  {ICLR} 2020, Addis Ababa, Ethiopia, April 26-30, 2020}. OpenReview.net.

\end{thebibliography}
\appendix
\appendix

\section{Statistics of the Pre-Training Data.}
\label{append:data}
We present dataset statistics for pre-training corpora in Table~\ref{tab:data_static}.

\section{Statics of Five Different Scenarios}
\label{append:scenarios}
We present dataset statistics for fine-tuning corpora in Table~\ref{tab:data_ft}.

\section{Detailed Ablation Experiments}
\label{append:ablations}
We show more detailed results of the ablation experiments on two language pairs in Table~\ref{tab:abla_exp}.

\section{Performance with Iterations for NAT}
\label{append:iter_NAT}
\begin{figure}[!h]
    \centering
    \includegraphics[scale=0.5]{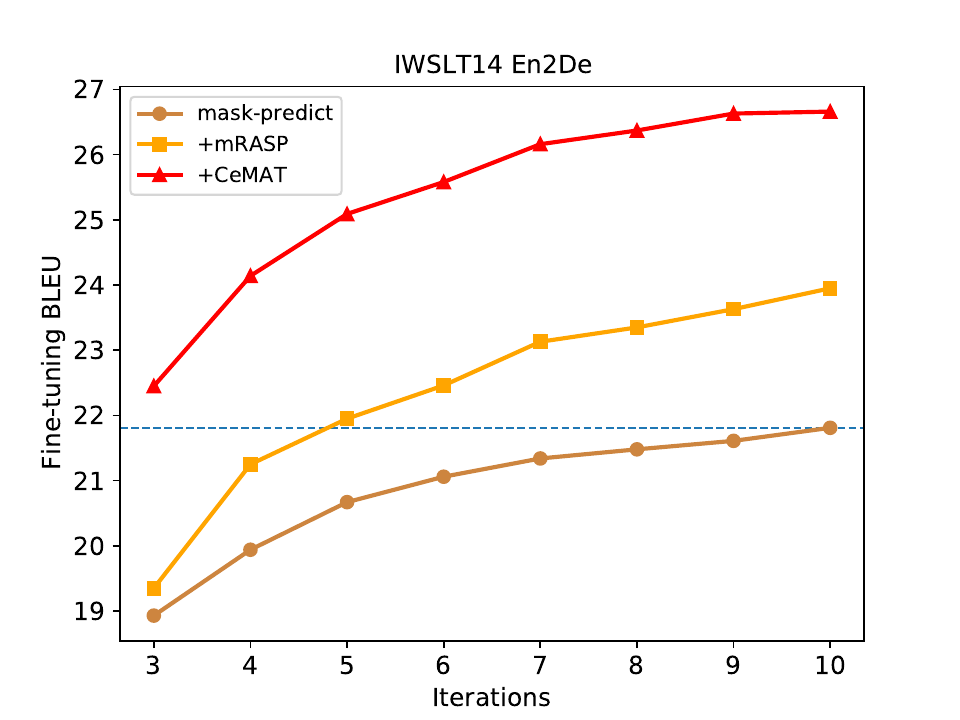}
    \caption{The performance of IWSLT14 En2De when decoding with different number of iterations.
    }
    \label{fig:iter1}
\end{figure}

\begin{figure}[!h]
    \centering
    \includegraphics[scale=0.5]{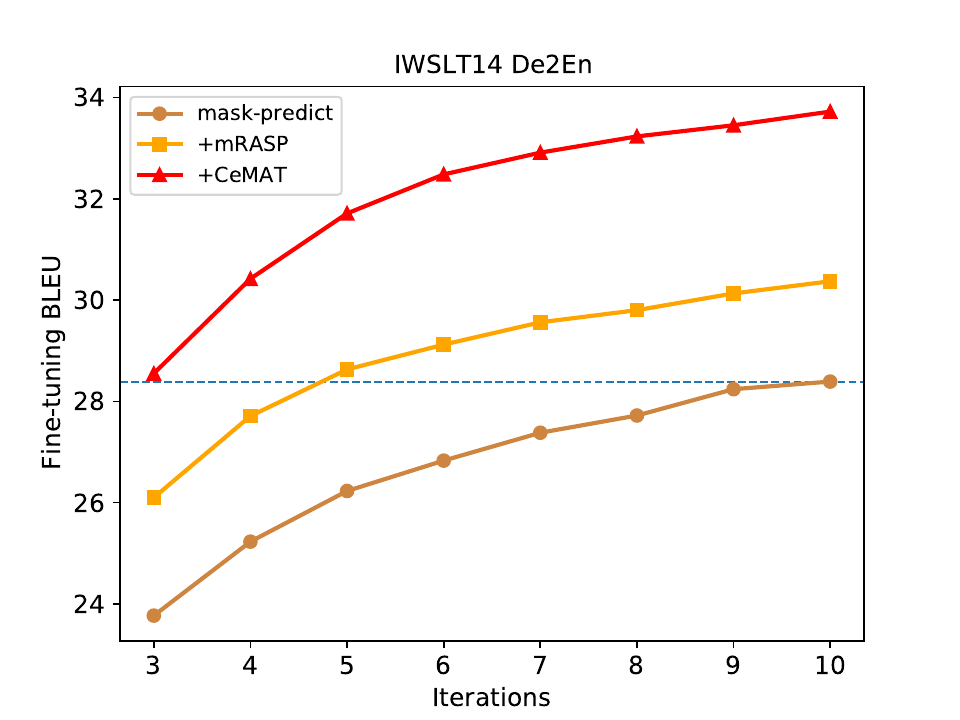}
    \caption{The performance of IWSLT14 De2En when decoding with different number of iterations
    }
    \label{fig:iter2}
\end{figure}

\begin{figure}[!hb]
    \centering
    \includegraphics[scale=0.5]{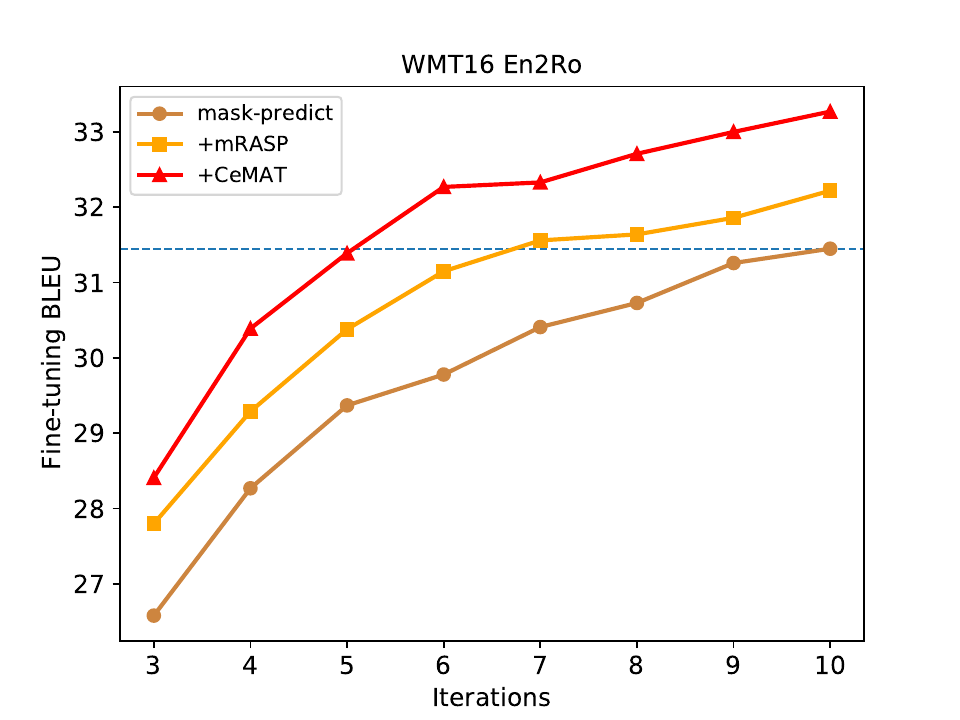}
    \caption{The performance of WMT16 En2Ro when decoding with different number of iterations.
    }
    \label{fig:iter3}
\end{figure}

\begin{figure}[!h]
    \centering
    \includegraphics[scale=0.5]{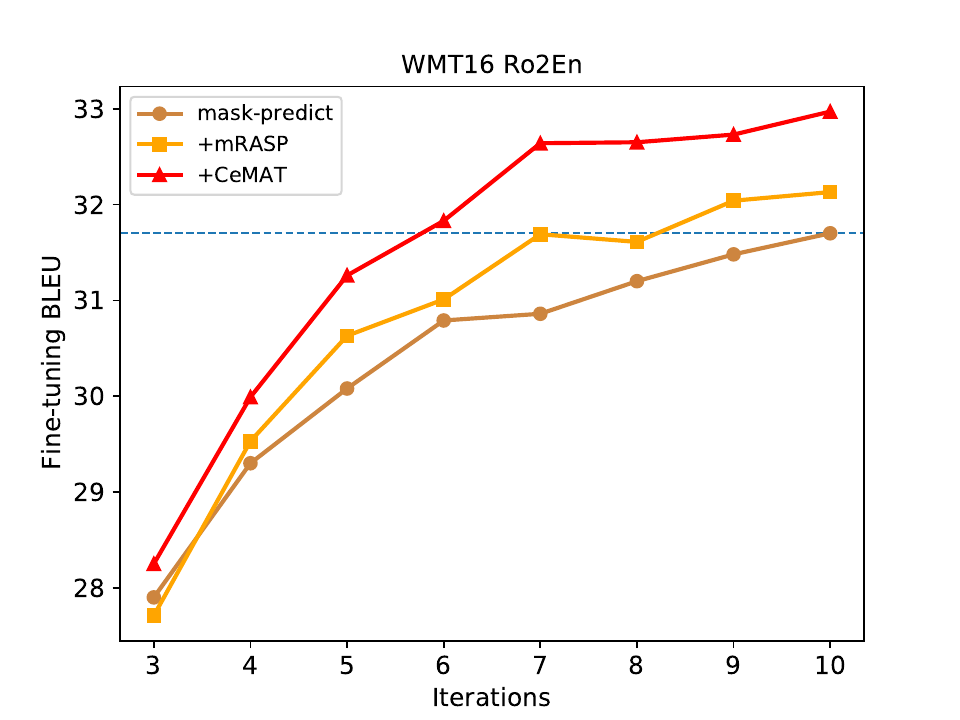}
    \caption{The performance of WMT16 Ro2En when decoding with different number of iterations.
    }
    \label{fig:iter4}
\end{figure}

\begin{figure}[!h]
    \centering
    \includegraphics[scale=0.5]{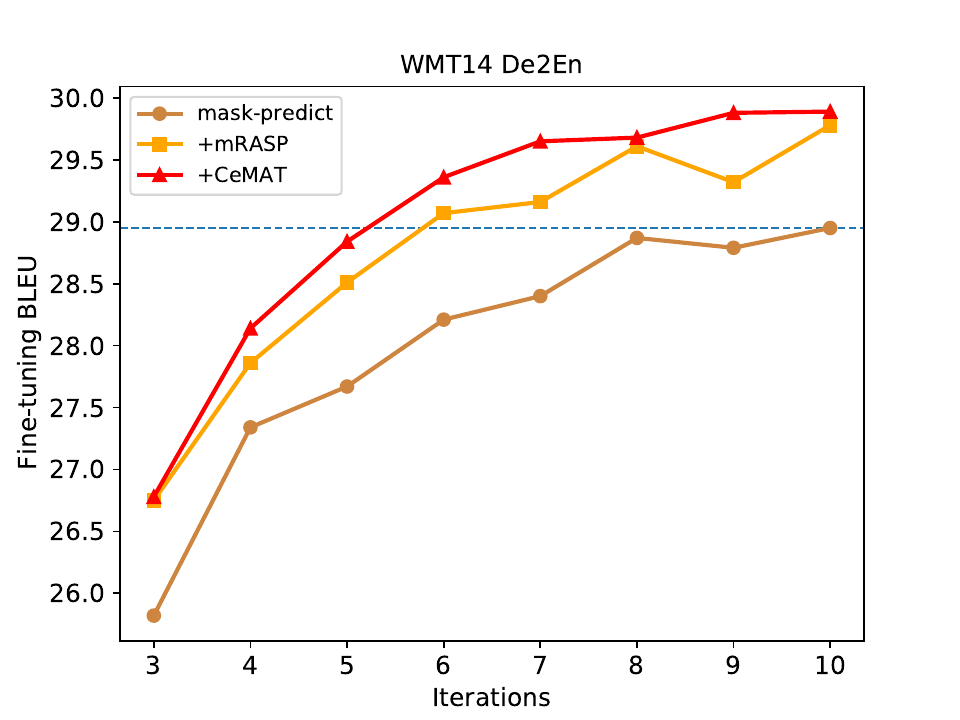}
    \caption{The performance of WMT14 De2En when decoding with different number of iterations.
    }
    \label{fig:iter5}
\end{figure}

\begin{figure*}[!htp]
    \centering
    \includegraphics[scale=0.5]{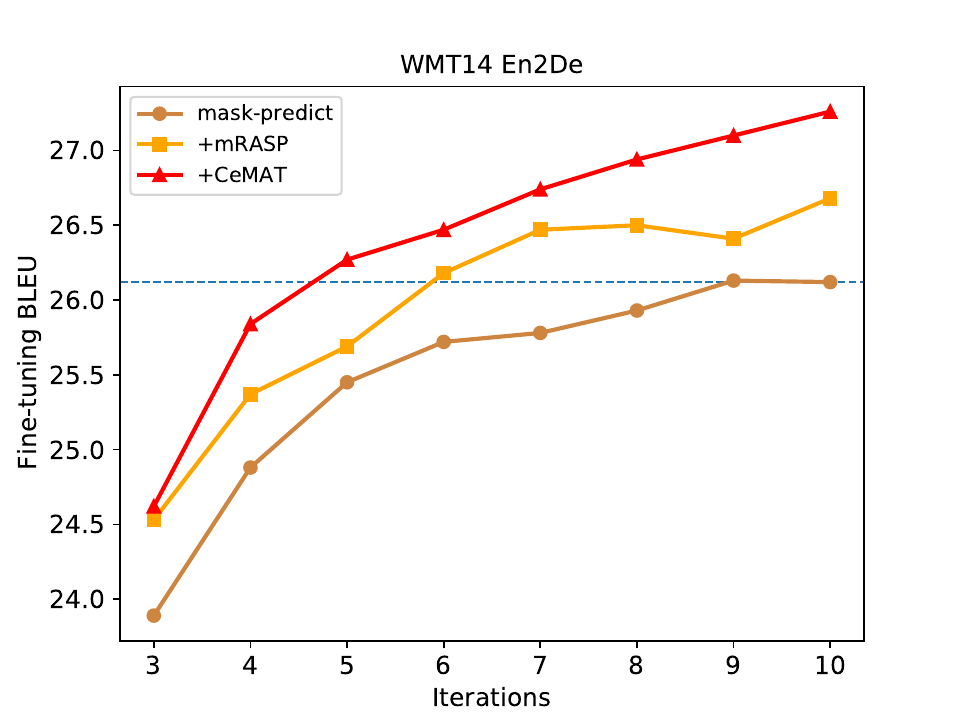}
    \caption{The performance of WMT14 En2De when decoding with different number of iterations.
    }
    \label{fig:iter6}
\end{figure*}

We present the dynamic performance on three language-pair datasets during iterative decoding in Figure~\ref{fig:iter1}, \ref{fig:iter2}, \ref{fig:iter3}, \ref{fig:iter4}, \ref{fig:iter5} and \ref{fig:iter6}.

\begin{table*}[!hbp]
\centering
    \begin{tabular}{l|lcc|l|lcc}
    \hline
    ISO & Language   &Bilingual  & Monolingual & ISO & Language   &Bilingual & Monolingual \\
    \midrule
    Gu  & Gujarati   & 11K         & 815K      & Ko  & Korean     & 1.4M        & --        \\
    Be  & Belarusian & 24K         & --        & Ms  & Malay      & 1.6M        & --        \\
    My  & Burmese    & 28K         & --        & Ru  & Russian    & 1.8M        & 9.9M      \\
    Mn  & Mongolian  & 28K         & --        & Fi  & Finnish    & 2M          & 9.9M      \\
    Af  & Afrikaans  & 40K         & --        & Ja  & Japanese   & 2M          & 3.4M      \\
    Eo  & Esperanto  & 66K         & --        & It  & Italian    & 2M          & 9.9M      \\
    Kk  & Kazakh     & 122K        & 1.8M      & Es  & Spanish    & 2.1M        & 9.9M      \\
    Sr  & Serbian    & 133K        & 3.7M      & Et  & Estonian   & 2.2M        & 5.3M      \\
    Mt  & Maltese    & 174K        & --        & Lt  & Lithuanian & 2.3M        & 2.8M      \\
    Ka  & Kannada    & 198K        & --        & Lv  & Latvian    & 3.0M        & 11.3M     \\
    He  & Hebrew     & 330K        & --        & Bg  & Bulgarian  & 3.1M        & 9.9M      \\
    Tr  & Turkish    & 383K        & 9.9M      & Vi  & Vietnamese & 3.1M        & --        \\
    Ro  & Romanian   & 770K        & 20M       & De  & German     & 4.6M        & 15M       \\
    Cs  & Czech      & 814K        & 9.9M      & Zh  & Chinese    & 21M         & 4.4M      \\
    Ar  & Arabic     & 1.2M        & --        & Fr  & French     & 36M         & 15M       \\
    El  & Greek      & 1.3M        & 8.3M      & En  & English    & --          & 15M       \\
    Hi  & Hindi      & 1.3M        & 9.9M      &     &            &             &          \\
    \hline
    \end{tabular}
    \caption{A list of 32 Enlish-centric language pair datasets. Among them, 21 languages have corresponding monolingual data. In this work, we using the ISO code represent the language name, and put them at the beginning of the source and target.}
    \label{tab:data_static}
\end{table*}

\begin{table*}[!htbp]
\centering
    \scalebox{1.0}{
    \begin{tabular}{llcc}
    \hline
    Lang-Pairs &  Source & Size  & Category       \\
    \midrule
    En-Kk      & WMT19  &97K   & low-resource        \\
    De-En      & IWSLT14 &159K & low-resource     \\
    En-Tr      & WMT17  &207K  & low-resource          \\
    En-Ro      & WMT16  &597K  & low-resource          \\
    En-Et      & WMT18  &1.9M & medium-resource        \\
    En-Fi      & WMT17  &2.7M & medium-resource       \\
    En-Lv      & WMT17  &4.5M  & medium-resource       \\
    En-De      & WMT14  &4.5M  & medium-resource       \\
    En-Cs      & WMT19  &11M   & high-resource         \\
    En-De      & WMT19  &38M   & extremely high-resource        \\
    En-Fr      & WMT14  &41M   & extremely high-resource         \\
    \hline
    \end{tabular}}
    \caption{The statistical information of the language pairs on \emph{low- / medium- / high- / extremely high-}resource for the machine translation task.
    }
    \label{tab:data_ft}
\end{table*}

\begin{table*}[]
\centering
    \begin{tabular}{lccccc}
    \hline
    Lang-Pairs                 & \multicolumn{2}{c}{Kk-En}  & \multicolumn{2}{c}{Et-En} &   Avg    \\
    Direction                  & $\rightarrow$                   & $\leftarrow$      & $\rightarrow$ & $\leftarrow$                   &    \\
    \hline
    w/  \emph{Bilingual}               & 7.8                     & 5.5                     & 24.4                    & 19.1                    & 14.2  \\
    w/  \emph{Monolingual}             & 5.4                     & 5.4                     & 23.5                    & 18.9                    & 13.3  \\
    w/  \emph{Bi- \& Monolingual} & 9.0                     & 5.6                     & 25.2                    & 19.0                    & 14.7  \\
    w/o \emph{Aligned CS masking}  & 8.4                     & 5.1                     & 24.3                    & 18.2                    & 14.0  \\
    w/o  \emph{Dynamic (masking:0.15)} & 7.3                     & 4.4                     & 23.5                    & 17.7                    & 13.2  \\
    w/o  \emph{Dynamic (masking:0.35)} & 8.8                     & 5.6                     & 23.7                    & 18.1                    & 14.1 \\
    \hline
    \end{tabular}
    \caption{Verification of the effectiveness of different techniques on two language pairs: Kk-En and Et-En. “w/ \emph{Bilingual}” denotes that we use only bilingual data when pre-training \ours; “w/ \emph{Monolingual}” denotes that we use only monolingual data when pre-training \ours; “w \emph{Bi- \& Monolingual}” denotes that when pre-training \ours, we use both bilingual and monolingual data; “w/o \emph{Aligned CS masking}” denotes that we pre-train \ours without \aligncs algorithm; “w/o \emph{Dynamic (masking:0.15)}” means that we use a fixed masking ratio with 0.15 for dual-masking; “w/o \emph{Dynamic (masking:0.35)}” means that we use a fixed masking ratio with 0.35 for dual-masking to make a more fair comparison with dynamic masking. To save computational resources, we use Transformer-base to obtain all the results of this experiment.}
    \label{tab:abla_exp}
\end{table*}

\end{document}